\documentclass[english]{amsart}
\usepackage[T1]{fontenc}
\usepackage[latin1]{inputenc}
\usepackage{geometry}
\usepackage{float}
\usepackage{mathrsfs}
\usepackage{amsmath}
\usepackage{amssymb}
\usepackage{graphicx}
\usepackage{subfigure}
\usepackage{multirow}
\usepackage{fancyheadings, lastpage}

\makeatletter

\floatstyle{ruled}
\newfloat{algorithm}{tbp}{loa}
\providecommand{\algorithmname}{Algorithm}
\floatname{algorithm}{\protect\algorithmname}

\usepackage{amsthm}

\usepackage{amsfonts}

\usepackage{epsfig}

\usepackage{bm}

\usepackage{mathrsfs}

\usepackage{enumerate}

\@ifundefined{definecolor}{\@ifundefined{definecolor}
 {\@ifundefined{definecolor}
 {\usepackage{color}}{}
}{}
}{}

\usepackage{subfig}\usepackage[all]{xy}

\newcommand{\bbR}{\mathbb R}
\newcommand{\bbE}{\mathbb E}

\newcommand{\cL}{\mathcal L}
\newcommand{\cX}{\mathcal X}
\newcommand{\cD}{\mathcal D}
\newcommand{\cC}{\mathcal C}
\newcommand{\cY}{\mathcal Y}
\newcommand{\cB}{\mathcal B}
\newcommand{\cA}{\mathcal A}
\newcommand{\cS}{\mathcal S}
\newcommand{\cN}{\mathcal N}

\newtheorem{rem}{Remark}[section]

\newcounter{hypA}

\newcounter{probP}

\usepackage{babel}\date{}

\author{David E. Bernholdt$^\$$}
\thanks{$^\$$Computer Science Department, 
Oak Ridge National Laboratory, Oak Ridge, TN, 37831}

\author{Mark R. Cianciosa$^*$}
\author{Clement Etienam$^\dagger$}
\author{David L. Green$^*$}
\author{Kody J. H. Law$^\dagger$}
\thanks{$^\dagger$School of Mathematics, University of Manchester, Manchester, UK, M13 4PL}

\author{J.~M.~Park$^*$}
\thanks{$^*$Fusion Department, Oak Ridge National Laboratory, Oak Ridge, TN, 37831}

\makeatother

\title[CCR method]
{Cluster, classify, regress: a general method for learning discontinuous functions}

\usepackage{babel}
\begin{document}

\begin{abstract}

This paper presents a method for solving the supervised learning problem in which the 
output is highly nonlinear and discontinuous. It is proposed
to solve this problem in three stages: 
(i) cluster the pairs of input-output data points, resulting in a label for each point; 
(ii) classify the data, where the corresponding label is the output; and finally
(iii) perform one separate regression for each class, 
where the training data corresponds to the subset of the
original input-output pairs which have that label according to the classifier.
%While these are the 3 fundamental building blocks of machine learning,
It has not yet been proposed to combine these 3 fundamental building blocks 
of machine learning in this simple and powerful fashion.
This can be viewed as a form of deep learning, where any of the intermediate 
layers can itself be deep.
The utility and robustness of the methodology is illustrated on some toy problems,
including one example problem arising from simulation of plasma fusion in a tokamak.

\end{abstract}

%+Title
%\begin{center}
\maketitle

\section{Introduction and motivation}\label{introduction}

%\section{Quantum Simulation Setup}\label{sec:quantum}

Modern times have seen an explosion of available data. This largely originated 
from the internet, although improved abilities to capture and store data have lead 
to similar data deluge phenomena in science and engineering applications.
Machine learning technology has recently achieved a level of maturity where it
is of increasing interest to adapt and apply the fundamental algorithms to 
problems in science and engineering. However, aside from the big data aspect, 
the nature of the problems which arise in science are often fundamentally 
different in character from the problems 
for which standard machine learning technology was developed. 
For example, many problems which arise in science and engineering involve
highly nonlinear and/or discontinuous functions over high-dimensional parameter spaces 
\cite{gardner2000construction, najm2009uncertainty, meneghini2017self}.
The term ``high-dimensional'' has different meanings in different contexts, 
but for the present work it will mean $\gtrsim 10$.  

The fantasy of the domain scientist or engineer is that there is a magical black box in 
which he can dump his data, and out of which will appear a miraculous machine capable 
of reproducing the results, to suitable accuracy, 
of his experiments or code which had been months or even generations in development. 
Indeed, sometimes this is possible with traditional existing methodology.
This problem has been studied for a long time in the approximation theory and statistics 
literature, where it may be referred to as a surrogate or an emulator 
\cite{sacks1989design, conti2010bayesian}.
The classical problems of machine learning are typically either of classification or regression type.
Highly nonlinear and discontinuous functions are something in between,
and their reconstruction has been the focus of much existing work in the literature 
\cite{eckhoff1995accurate, adalsteinsson1995fast, wang2003level, gelb2002spectral, archibald2005polynomial, archibald2009discontinuity, pfluger2010spatially, jakeman2011characterization}, 
but it is still an active area \cite{batenkov2015complete, gorodetsky2014efficient, zhang2016hyperspherical, monterrubio2018posterior, dunlop2017hierarchical}. 
In the present work, we propose a simple general framework 
which can be utilized to solve such problems, using as components the arsenal of available 
machine learning algorithms \cite{murphy2012machine, bishop2006pattern, friedman2001elements, rasmussen2006gaussian, goodfellow2016deep}.

The paper will be organized as follows. In Section \ref{sec:mathsetup}, 
the mathematical problem and algorithm will be presented. 
Section \ref{sec:active} describes an adaptive active learning extension 
to the current algorithm. The approach is capable of either selecting an 
opimal subset of training data among an existing set of labeled data, 
or adaptively selecting training data to label from the domain of all possible 
(unlabeled) data.
Section \ref{sec:numerics} will feature some numerical illustrations. 
%{\bf and comparison with existing algorithms???}.
Finally Section \ref{sec:conclusions} will conclude 
and discuss some extensions of the present work.

\section{Mathematical Setup and Algorithm}\label{sec:mathsetup}

The point of departure is a set of labeled training data $\{(x_i,y_i)\}_{i=1}^N$, 
where $(x_i,y_i) \in \cX\times \cY$ 
are assumed to correspond to input and output of some model
$$
y_i \approx f(x_i) \, ,
$$
and are otherwise considered independent. 
In other words, we want to do supervised learning. 
It is furthermore assumed that the model $f: \cX \rightarrow \cY$ is highly irregular, 
including sharp features arising from strong nonlinearities, 
and most notably including discontinuities. 
The output space is taken as $\cY=\bbR$ for simplicity. 
The important point is that it is continuous.
The results can be easily generalized to vector valued outputs.
We will also assume $\cX = \bbR^d$, although this can also be relaxed.
If the data are exact, one can utilize interpolation or projection methods 
\cite{bungartz2004sparse, xiu2010numerical}, which are generally outside the repertoire of machine learning.
However, even in the case of exact data, these methods can be particularly 
sensitive to the choice of input data points used in the approximation, and can 
yield poor results. 
Ordinarily one would look to regression methods in this context, 
since the output space is continuous. 
Regression methods handle noisy data elegantly, and are also more robust to 
the peculiarities of the input data. 
However, typically regression methods 
will fail miserably when $f$ has discontinuities (as will interpolation and projection methods). 
It is therefore natural to attempt to %remove the discontinuities 
partition the domain according to the discontinuities, and
then perform piecewise regression on the continuous components.
This strategy has been proposed before in the literature 
\cite{eckhoff1995accurate, adalsteinsson1995fast, wang2003level, gelb2002spectral, gorodetsky2014efficient, dunlop2017hierarchical, zhang2016hyperspherical}.
%however typically the focus is on finding the discontinuity itself, and 
However, a point which remains unclear is how to identify the continuous components, or equivalently the discontinuous hyper-surfaces, %label the data, 
and importantly how to do so without relying on any grid of the input space.
Here we propose to use a clustering algorithm to label the input-output 
training data pairs, 
coupled with a classification algorithm to label new inputs.
To be precise the full method in generality is as follows.

\noindent{\bf (I) Cluster.} First seek a label function which clusters with 
the training input-output pairs as its inputs 
\begin{equation}\label{eq:label}
\lambda : \cX \times \cY \rightarrow \cL := \{1, \dots, L\} \, .
\end{equation}
%In certain cases it may be possible to simply label the data
The label function typically minimizes an objective function of the form
\begin{equation}\label{eq:clust}
\Phi_{\rm clust}(\lambda) = \sum_{l=1}^L \sum_{i \in S_l} \ell_l(x_i,y_i) \, ,
\end{equation}
where $S_l = \{ (x_i,y_i) ; \lambda(x_i,y_i) = l \}$, 
and $\ell_l$ is some loss function associated to cluster $l$. 
For example, letting $z_i=(x_i,y_i)$, $\ell_l = |  z_i -  \mu_l |^2$, 
and $\mu_l = \frac1{|S_l|}\sum_{i \in S_l} z_i$, 
where $|\cdot|$ denotes the Euclidean norm 
for points and counting norm for sets,
we have $K-$means clustering.
Choosing $L$, for example using the elbow method, 
completes the clustering phase of the algorithm 
\cite{bishop2006pattern, friedman2001elements}.

\noindent{\bf (II) Classify.} Letting $l_i = \lambda(x_i,y_i)$, 
we now have an expanded set of training data $\{(x_i,y_i,l_i)\}_{i=1}^N$.
The classification phase proceeds to find a function which appropriately labels the inputs 
according to the labels identified in the cluster phase:
 \begin{equation}\label{eq:class}
f_c : \cX \rightarrow \cL \, .
\end{equation}
The classifier could be non-parametric or parametric, but the important point 
is that for each $x \in \cX$ it provides an estimate $f_c : x \mapsto f(x) \in \cL$
such that $f_c(x_i) = l_i$ for the majority of the data. 
This is crucial for the ultimate fidelity of the prediction.
Note that the output data $\{y_i\}$ is ignored in this phase.
The classification function minimizes 
\begin{equation}\label{eq:class}
\Phi_{\rm class}(f_c) = \sum_{i=1}^N \phi_c(l_i, f_c(x_i)) \, ,
\end{equation}
where $\phi_c : \cL \times \cL \rightarrow \bbR_+$ 
is small if $f_c(x_i) = l_i$ and otherwise is not.
For example, we can choose $f_c(x) = {\rm argmax}_{l \in \cL} g_l(x)$,
where %$g(x) = [g_1(x), \dots, g_L(x)]$ with 
$g_l(x)>0$, $\sum_{l=1}^L g_l(x)=1$ is a soft classifier,
and $\phi_c(l,f_c(x)) = - \log(g_l(x))$ 
\cite{bishop2006pattern, friedman2001elements, murphy2012machine}, 
corresponding to cross-entropic loss.

\noindent{\bf (III) Regress.} The final phase of approximation is to find a function
which appropriately identifies the original output given the input and the label 
 \begin{equation}\label{eq:regress}
f_r : \cX \times \cL \rightarrow \cY \, .
\end{equation}
The regressor could be non-parametric or parametric, but the important point 
is that for each $(x,l) \in \cX \times \cL$ it provides an estimate 
$f_r : (x,l) \mapsto f_r(x,l) \in \cY$ such that now 
$f_r(x,f_c(x)) \approx y$ for both training data \emph{and test data}.
If successful then we can expect good reconstruction error for this ultimate predictor
\begin{equation}\label{eq:machine}
f : \cX \rightarrow \cY \, ,
\end{equation}
where $f(\cdot) = f_r(\cdot , f_c(\cdot))$.
The regression function can be found by minimizing
\begin{equation}\label{eq:reg}
\Phi_{r}(f_r) = \sum_{i=1}^N \phi_r(y_i, f_r(x_i, f_c(x_i))) \, ,
\end{equation}
where $\phi_r : \cY \times \cY \rightarrow \bbR_+$ 
minimized when $f_r(x_i, f_c(x_i)) = y_i$ .
In this case we can choose 
$\phi_r(y, f_r(x, f_c(x))) = | y - f_r(x, f_c(x))|^2$, 
\cite{bishop2006pattern, friedman2001elements}.
Note that it is computationally more expedient to partition the data
into $C_l = \{ i ; f_c(x_i) = l \}$, for $l=1,\dots, L$, 
and then perform $L$ separate regressions (which can also be done in parallel)
\begin{equation}\label{eq:regl}
\Phi_{r}^l(f_r(\cdot, l)) = \sum_{i \in C_l } \phi_r(y_i, f_r(x_i, l)) \, .
\end{equation}

%This will be given as the composition of two functions. 
%The first of which is the classifier above,
%and the second is a regressor which maps inputs and labels to outputs

{\rem A practical consideration with the proposed methodology 
is the relative scaling of the data in $x = (x^1,\dots, x^d) \in \bbR^d$ and $y \in \bbR$. 
We have $|(x,y) - (x',y')|^2 = (y-y')^2 + | x - x' |^2$.
In particular, notice that if the range is a small fraction of the 
size of the domain, then even large discontinuities in the output
may not be revealed in the clustering. 
Therefore, it is recommended for the clustering to scale as follows
\begin{eqnarray}\label{eq:scaling}
\tilde{x}^j &=& (x^j - {\rm min}_{i\in\{1,\dots, N\}} x_i^j)/|(x^j - {\rm min}_{i\in\{1,\dots, N\}} x_i^j)| \, \quad {\rm for}\quad j=1,\dots, d, \\
\tilde{y} &=& C (y - {\rm min}_{i\in\{1,\dots, N\}} y)/|(y - {\rm min}_{i\in\{1,\dots, N\}} y)| \, ,
\end{eqnarray}
for $C>1$.
In particular, it is recommended to choose $C=10d$, where
$d={\rm dim} x$.
One could also use some other type of standardization, e.g. 
subtracting the mean and dividing by the standard deviation.
For the regression it is recommended to set $C=1$ above, 
as smaller variations in $y$ are easier to deal with for regression.}

{\rem A critique of this method is that it re-uses the data in each phase. 
We note that %once phase (I) has been completed, 
there is a Bayesian formulation, which will be considered in a future work.
%of the final two phases (II-III). 
Let $D = \{(x_i,y_i)\}_{i=1}^N$, 
assume we have parametric models for the classifier $g_l(\cdot ; \theta_c) = g_l(\cdot ; \theta_c^l)$, 
and the regressor $f_r(\cdot , l ; \theta_r^l)$, 
for $l=1,\dots, L$, where $\theta_c = (\theta_c^1, \dots, \theta_c^L)$ and
$\theta_r = (\theta_r^1, \dots, \theta_r^L)$ index the parameters corresponding
to each class, and let $\theta = (\theta_c, \theta_r)$. 
Then the posterior density has the form
$$
\pi( \theta, l | D) \propto 
\prod_{i=1}^N \pi(y_i | x_i, \theta_r, l) \pi(l | x_i, \theta_c) \pi(\theta_r) \pi(\theta_c) \, ,
$$
where, for example,
$$
\pi(y_i | x_i, \theta_r, l) \propto \exp( -\frac12 | y_i - f_r(x_i, l; \theta_r^l) |^2 ) \, ,
$$
and
$$
\pi(l | x_i , \theta_c) = g_l(x_i ; \theta_c) \, .
$$
In this case we may take for example 
$$
g_l(x ; \theta_c) = \frac{\exp(h_l(x ; \theta_c^l))}{\sum_{l=1}^L \exp(h_l(x ; \theta_c^l))} \, , 
$$
where $h_l(\cdot ; \theta_c^l)$ are some standard parametric regressors.

It is important to 
note that fully Bayesian solutions, while elegant, clean, and complete
with Occam's razor, are also very expensive. 
This example is grid-free, at least, 
and this feature will be desirable for high-dimensional problems of the type considered here. 
It will be interesting to compare this method 
to grid-based Bayesian approaches such as
\cite{dunlop2017hierarchical, monterrubio2018posterior}.
If the regressors are defined in terms of grids, 
for example $\theta_r^l$ are the coefficients of expansion in 
piecewise linear finite element nodal basis functions \cite{zienkiewicz1977finite},  
then there are similarities between the methods.}

{\rem Another important critique is that the method will struggle if there are discontinuities which appear or vanish within the domain, for example the product of a heaviside and a vanishing function,
as the phase (I) clustering will fail in this case. 
However, in practice, we have found that the method is able to 
handle such cases with a very large number of clusters, 
i.e. much greater than the number of continuous components. }

\section{Active learning extension}\label{sec:active}

Here it is described how to embed the algorithm above into 
an active learning strategy \cite{settles2009active}. 
First, observe that the fundamental 
steps above are (I-II), as a continuous approximation (III) 
of a function with a discontinuity will always be a poor approximation.
Indeed the error in such approximations is typically 
concentrated around the discontinuities.
For this reason, it is preferable to choose a soft-classifier (or probabilistic classifier) \cite{murphy2012machine}, 
even if the ultimate classification is taken as a thresholding of such continuous function,
as described above in (II) for multiclass logistic regression.
Suppose we have a soft classifier $\{g_l(x)\}_{l \in \cL}$, $\sum_{l\in \cL} g_l(x)=1$,
such that $g_l(x)$ represents the probability that point $x$ is in class $l$, and 
our classification is given by $f_c(x) = {\rm argmax}_{l \in \cL} g_l(x)$. 
Suppose we have some initial set of $N_{\rm init}$ training data points
$\cD_{\rm init} := \{x_{-N_{\rm init}+1},x_{-N_{\rm init}+2}, \dots, x_0\}$.

{\bf Active Strategy 1.}
It is natural to seek, for $n>0$,  
\begin{equation}\label{eq:iteration}
x_n = {\rm argmin}_{x \in \cD_n} {\rm max}_{l \in \cL} g_l(x) \, ,
\end{equation}
for some appropriate compact domain $\cD_n \subseteq \cX$. 

\noindent {\bf Active Strategy 1a.}
An extremely simple example would be the case in which 
$\cD_1 = \cD_{\rm init}$ and 
$\cD_n = \{x_1, \dots x_{N_{\rm res}}\} \backslash \cD_{n-1}$ is a finite subset of the domain,
consisting of $N_{\rm res}$ points which we refer to as the ``reservoir'',
which may for example be existing labeled points which we are aiming to 
scrupulously include in our training algorithm. 

\noindent {\bf Active Strategy 1b.}
Alternatively, we may constrain the algorithm to search only some continuous subset 
of the possible (currently unlabelled) input data. 
Here we either need an a priori defined domain, or some sensible 
way of adapting the domain. If we know the 
input data of interest lies within some hypercube $\cD=\prod_{i=1}^d [a_i,b_i]$,
then it may be reasonable to choose this as $\cD_n=\cD$.
However, one expects that often the data may be concentrated on some 
manifold within such a hypercube. In that case, the naive strategy just mentioned
would lead to many points in the large volume of the parameter space which 
is uninteresting and will never be queried in practice. 
An alternative which may be viable in the case that the initial data 
$\cD_{\rm init}$ is suitably rich, and concentrated on the appropriate 
data manifold, would be to let $\cD = {\rm conv} \cD_{\rm init}$, 
the convex hull of the initial data set. 
Note that in this case the training data will always remain in this set.

{\bf Active Strategy 2.}
This strategy operates on the assumption of the latter version of 
strategy 1b, i.e. that $\cD = {\rm conv} \cD_{\rm init}$ provides suitable coverage of the 
input domain of interest. Let $\cD_1 = \cD_{\rm init}$.
Let $\cN^n_k(x)$ denote the $k$ nearest neighbors of $x$ in $\cD_n$, where
$k\geq 1$. Let 
$$
\cS_n = \{x \in \cD_n ; \exists y \in \cC_n(x)\} \, , \qquad 
\cC_n(x) = \{ y \in \cN_k(x) ;  f_c(x) \neq f_c(y) \} \, .
$$
Now, let the batch of new training data at step $n$, $\cB_n$, 
be defined by those points $z^*$, one for each $x\in \cS_n$ 
and each $y\in \cC_n(x)$, such that
%$$
%\cB_n = \{ z \in \cD ; 
$$
z^* =  {\rm argmin}_{z \in \cA} {\rm max}_{l \in \cL} g_l(z), \quad 
\cA=\{ z\in \cD; z = \lambda x + (1-\lambda) y\} %, ~~ y = {\rm argmin}_{y \in \cN_k(x)} |y-x| \}  
\, .
$$
This strategy is prone to degeneration, and so has to be regenerated once in a while, 
for example with strategy 1b or strategy 3 below.
When all $x\in \cS_n$ and $y\in \cC_n(x)$ are exhausted, we have 
$n^*\sum_{x \in \cS_n} |\cC_n(x)|$ new data points, $\cD^*$
which are concatenated to the present set $\cD_{n+1} = \cD_n \cup \cD^*$.
 
 {\bf Active Strategy 3.}
 Here we choose some set $\cA$ (different from above), 
 for example with any one of the strategies above, 
 and we then choose points $z_i(x)\sim Q(x,\cdot)$, for $i=1,\dots, n$, 
 from a Markov kernel $Q$ (i.e. $Q(x,\cdot)$ is a probability distribution for 
 each $x\in \cX$), for each $x\in \cA$. 
 We could let $Q(x,\cdot)$ be a uniform on the 
 $d-$dimensional hypercube centered at $x$, or a 
 normal centered at $x$. 
 It can be uniform/standard, or the covariance could be determined 
 by the sample covariance of $\cN^n_k(x)$ with a suitably large $k$,
 in order to stay faithful to the local manifold structure of the data.

{\rem Note with strategy 1 that there are likely infinitely many solutions 
to the optimization problem. Consider binary linear classification, where 
we choose label $1$ if $g_1(x)>1/2$ and label $2$ otherwise, and there 
is a separating hyperplane if $g_1$ is linear. 
Along the hyperplane we have $g_1(x)=1/2$,
i.e. there is a linear subspace of dimension $d-1$ which solves the problem
\eqref{eq:iteration}. This is not a major concern, 
as we only need one point, 
and any of these points will equivalently enrich the training set.}

{\rem We are mostly concerned with
the case in which we can label any point, but the labelling itself is the limiting 
cost. So active strategy 1a may be of limited practical value.}

{\rem The choice of domain $\cD$ is important for strategy 1b. 
If $\cD = \cX = \bbR^d$, then the strategy above is likely to seek
points outside the convex hull of the existing training data, where
nothing is known about the classifier. 
The danger is that we may not want to know 
anything about the classifier in that region.}

{\rem Motivated by the discussion at the beginning of this section, 
all strategies presented here are based on what is known as 
\emph{uncertainty sampling} \cite{settles2009active}. 
Other objectives can be used as well (or none at all, i.e. random sampling), 
including
\begin{enumerate}
  \item Entropy sampling: one chooses the point whose class probability has the largest entropy.
  \item Margin sampling: one chooses the point for which the difference between the most and second most likely classes is the smallest.
\end{enumerate}}

{\rem This section is not to be confused with simple adaptive online learning, 
in which the machine may be embedded within a workflow and queried regularly
for some input $x_{\rm new}$.
In this case, one would employ a similar but distinct approach.
First, evaluate the fitness of the machine for the queried point 
$x_{\rm new}$. For example, as in \eqref{eq:iteration} we may compute 
${\rm max}_{l \in \cL} g_l(x_{\rm new})$. 
Then, if the fitness is suitable (in this case the probability of the most likely 
class being sufficiently close to 1), we proceed with the machine.
If the fitness is low, we augment the training set with this new point 
and refine the machine. In this case, 
one may venture to decide if the new query point is suitably close to the existing
set of training data, and if not then refine within the convex hull as described in 
strategy 1b or 2 above.
}

\section{Method used and Numerical examples}\label{sec:numerics}
%\subsection{Clustering Scheme} 

Clearly there is enormous potential for exploring various configurations of the component algorithms, but since the purpose of this paper is to introduce the method and illustrate its power on some simple examples we choose only some of the most basic component algorithms.

For clustering, K-means clustering \cite{bishop2006pattern} is used, 
as described in Section \ref{sec:mathsetup} (I).
The standard iterative algorithm is used to optimize the objective function \eqref{eq:clust}.
The elbow method is used to determine the value of $L$ \cite{bishop2006pattern}.
%\noindent 
%
%\noindent 
%\subsection{Regression/Classification scheme}
For regression and classification, multi-layered perceptrons (MLPs) are used \cite{bishop2006pattern},
with a single hidden layer of $100 d$ neurons and ReLU$(x)={\rm max}\{0,x\}$ 
activation function for both. 
For the multi-classification, the output layer is given by a softmax, 
with a cross-entropy loss function, as described in Section \ref{sec:mathsetup} (II).
For regression, the output activation is linear, and the loss function is quadratic, 
as described in Section \ref{sec:mathsetup} (III).  
A quadratic regularization is used in both cases, with parameter $0.001$. 
The adaptive stochastic gradient descent solver Adam \cite{kingma2014adam}
is used to optimize \eqref{eq:class} and the 
$L$ functions \eqref{eq:regl}. 
A validation fraction of $0.1$ is employed and the 
number of iterations are limited by a maximum of $200$.
In examples 3 and 4, a random forest method \cite{breiman2001random} is used, 
in which $100$ bootstrapped classification or regression 
decision trees are aggregated \cite{breiman1996bagging},
where each one involves a subset of only ceil$(d/3)$ input parameters.
The scikit learn python package \cite{pedregosa2011scikit} is used 
for all the basic learning algorithms employed.

\subsection{Numerical experiments}

\begin{figure}%[!htb]
%\begin{center}
\centering
\subfigure[Numerical example 1, $f_1$.]{
%\label{fig:one}
\includegraphics[width=5.27in, height=1.30in, keepaspectratio=false]{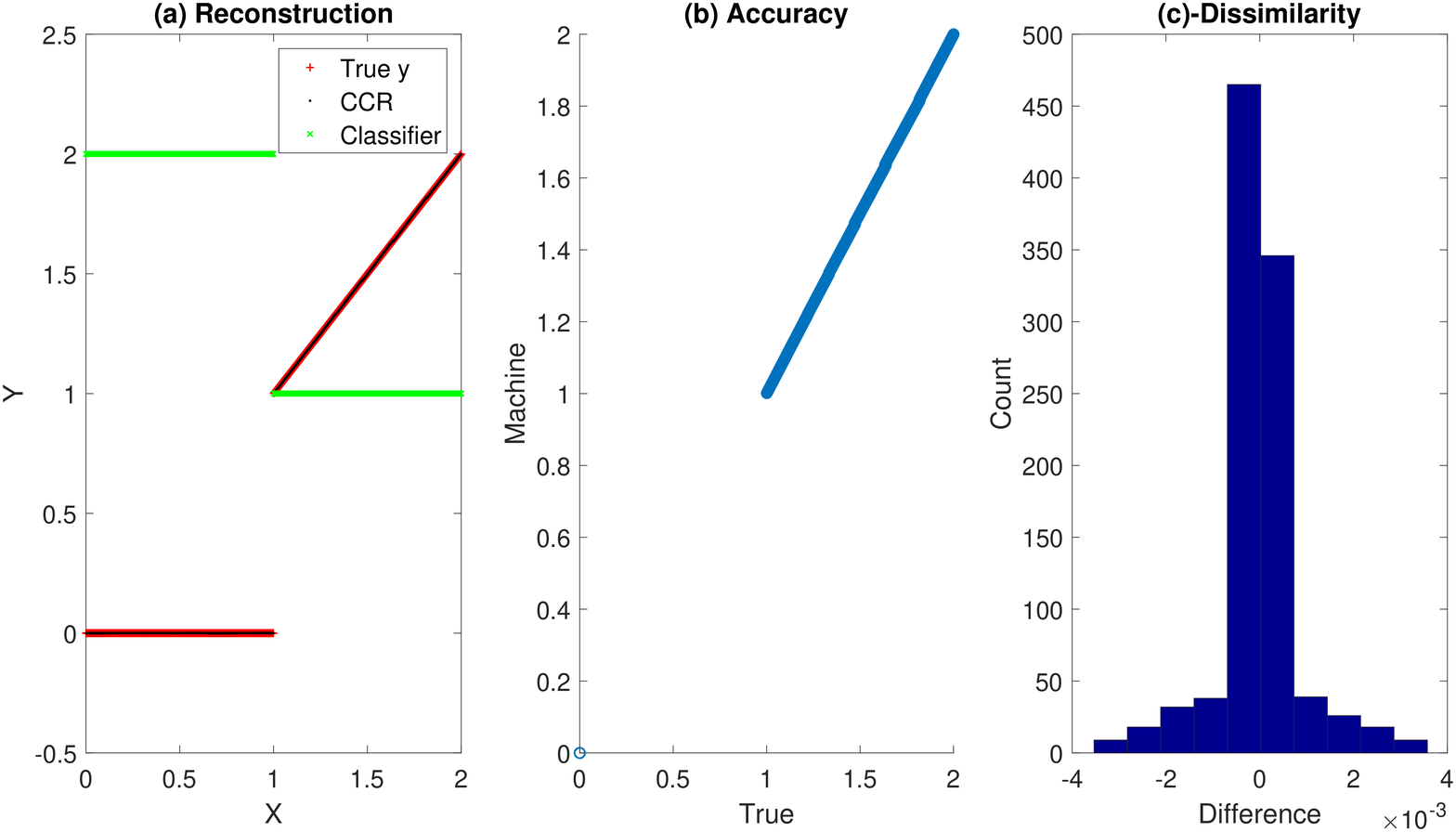}}
%\includegraphics[width=2.6in, height=1.5in, keepaspectratio=false]
%{NUM1.eps}}
%\[center]{Figure 1:}
%\end{center}
%\caption{Numerical example 1. 
%The function is plotted in Panel (a), 
%along with the clustering values of the training data.
%Panel (b) shows prediction results on test data: 
%the final CCR machine output $f_r(x ; f_c(x))$, 
%the true $y(x)$, and the intermediate $f_c(x)$.
%Panel (c) shows a scatter plot of the true $y(x)$ and the CCR machine $f_r(x ; f_c(x))$,
%illustrating the correlation.
%Panel (d) shows a histogram of $f_r(x ; f_c(x))-y(x)$,
%illustrating the dissimilarity between the CCR reconstruction and the truth.}
%\label{fig:2a}} 
%\end{figure}
%\end{subfigure}
%\begin{figure}[!htb]
%\begin{center}\label{fig:two}
%\centering
%\qquad
\subfigure[Numerical example 2, $f_2$.]{
%}[t]{0.5\textwidth}
%\label{fig:two}
\includegraphics[width=5.27in, height=1.30in, keepaspectratio=false]{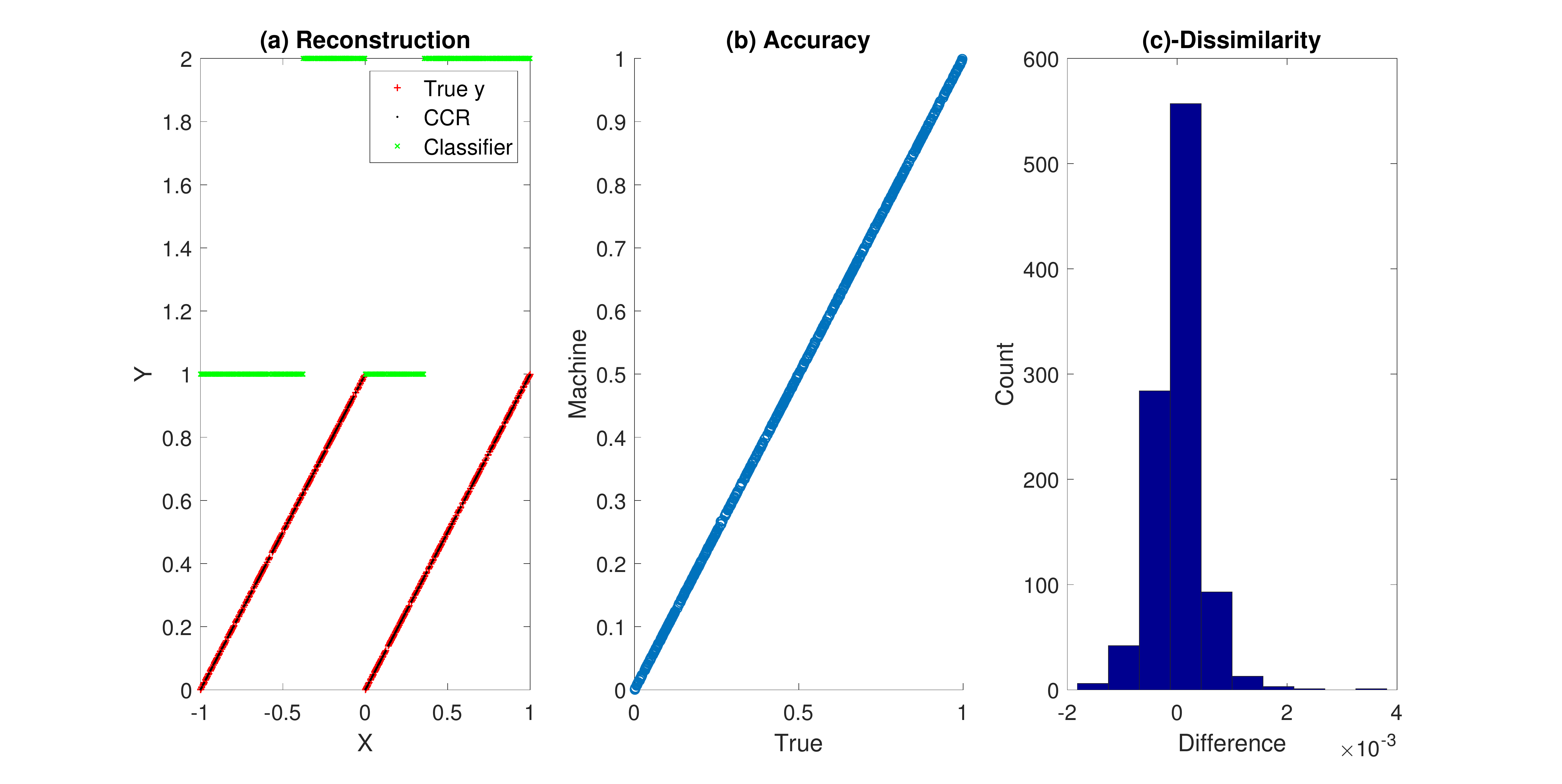}}
\subfigure[Numerical example 3, $f_3$.]{
\includegraphics[width=5.27in, height=1.3in, keepaspectratio=false]{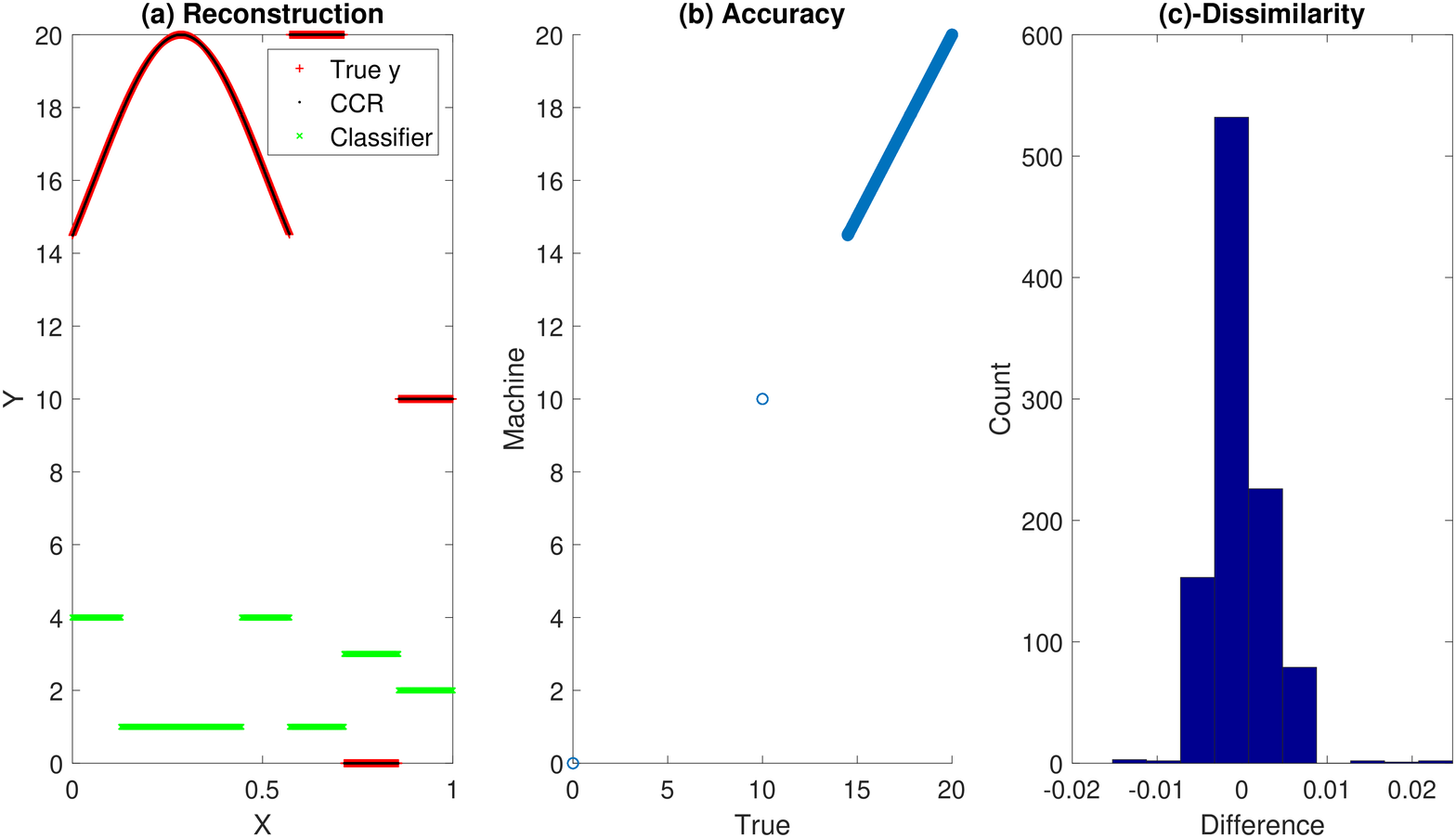}}
%\includegraphics[width=2.6in, height=1.5in, keepaspectratio=false]
%{NUM2.eps}}
%\center \textbf{Figure 2:}
%\label{fig:2a}} 
%\end{center}
%\caption{Numerical example 2. 
%The function is plotted in Panel (a), 
%along with the clustering values of the training data.
%Panel (b) shows prediction results on test data: 
%the final CCR machine output $f_r(x ; f_c(x))$, 
%the true $y(x)$, and the intermediate $f_c(x)$.
%Panel (c) shows a scatter plot of the true $y(x)$ and the CCR machine $f_r(x ; f_c(x))$,
%illustrating the correlation.
%Panel (d) shows a histogram of $f_r(x ; f_c(x))-y(x)$,
%illustrating the dissimilarity between the CCR reconstruction and the truth.}
\caption{Numerical examples 1 (row a), 2 (row b), and 3 (row c). 
The functions are plotted in columns (a), 
along with the final CCR machine output $f_r(x ; f_c(x))$, 
%the true $y(x)$, 
and the intermediate $f_c(x)$.
%clustering 
%classification values. %of the training data.
Columns (b) 
%show prediction results: %on test data: 
%the final CCR machine output $f_r(x ; f_c(x))$, 
%the true $y(x)$, and the intermediate $f_c(x)$.
%columns (c) 
show a scatter plot of the true $f(x)$ and the CCR machine $f_r(x ; f_c(x))$,
illustrating the correlation.
Panel (d) shows a histogram of $f_r(x ; f_c(x))-f(x)$,
illustrating the dissimilarity between the CCR reconstruction and the truth.}
%\end{subfigure}
\label{fig:one}
\end{figure}

In this section several prototype models are considered.
First, let $\cX=\bbR$ and consider the simple example 
%of the product of the 
%heaviside function with a linear function 
$f_1(x) = x{\bf 1}_{x\geq 1}$, 
%where 
%%$H: \bbR \rightarrow \{0,1\}$ is defined as 
%\begin{equation}\label{eq:heavy}
%H(x) = {\bf 1}_{x\geq 0} \, ,
%\end{equation}
where 
\[
{\bf 1}_{A}(x)=
\left \{ \begin{array}{cc} 1 & x\in A \, \\
0 & {\rm else} \\
\end{array} \right \} \, .\]
The function is plotted in Figure \ref{fig:one} row (a), column (a), 
along with the prediction results of the final CCR machine output 
$f_r(x ; f_c(x))$, and the intermediate $f_c(x)$.
%and classification. %clustering values of the training data.
%Figure \ref{fig:one}a(b) shows prediction results on test data: 
%the final CCR machine output $f_r(x ; f_c(x))$, 
%the true $y(x)$, and the intermediate $f_c(x)$.
Figure \ref{fig:one} row (a), column (b) shows a scatter plot of the true 
$f_1(x)$ and the CCR machine $f_r(x ; f_c(x))$, illustrating the correlation.
Figure \ref{fig:one} row (a), column (d) shows a histogram of $f_r(x ; f_c(x))-y(x)$,
illustrating the dissimilarity between the CCR reconstruction and the truth.

%\afterpage{\clearpage}
%\noindent \includegraphics*[width=4.27in, height=1.95in, keepaspectratio=false]{NUM1}
%\noindent \textbf{Figure 1:}
Notice that the previous example could have been dealt 
with using a simple clustering of the output values $y \in \bbR$ alone.
Consider the slightly more complicated function
$f_2(x) = (x+1) {\bf 1}_{x <0} + x{\bf 1}_{x\geq 0}$, 
for $\cX = [-1,1]$. 
Here the $y$-values alone cannot be used for clustering,
%In fact, here $y$ values alone 
as they suggest a single cluster is optimal 
(e.g. using the elbow method would return $L=1$).
However, our method of clustering input-output pairs $(x,y)\in \bbR^2$
works here, and elbow returns $L=2$. %beautifully. 
The method is illustrated on this example in Figure %\ref{fig:two},
\ref{fig:one}, row (b)
whose panels are the same as Figure \ref{fig:one}a.
Note that due to the gross simplicity of the model, 
if we look for 2 clusters based on $y$ values alone, 
then we would get something like $y\in[0,0.5]$ 
and $y\in (0.5,1]$. 
In other words, we would bypass the issue of the discontinuity, 
as a single class would not span the discontinuity, 
and this is the primary requirement in order for the 
piecewise regression to perform well. 
This will not hold in general, 
but it is illustrative of the robustness afforded by choosing extra clusters
and extra amplification of $y$.
%This is indicative that more clusters improves the robustness of the method.

%The function is plotted in Figure 2(a), 
%along with some noisy observations, 
%and the ultimate reconstruction. 
%Figure 2(b), 2(c) and 2(d) show, respectively, 
%the clustering result with our novel CCR method, the r-fit correlation line between CCR and the true signal and finally a histogram showing the dissimilarity between the signal recovered with CCR and the true signal

Note that the first 2 functions can both be reconstructed exactly with relatively simple
multilayer perceptrons, \emph{provided discontinuous activation functions are utilized}. 
Consider a 3 node hidden layer with $z_1 = {\rm max}\{0,x\}$, $z_2={\rm max}\{0,-x\}$, 
$z_3={\bf 1}_{x\geq 0}$, and $z_4={\bf 1}_{-x\geq 0}$. 
Then we have $f_1(x) = 1+ z_1 - z_4$ and $f_2(x) = 1- z_2 + z_1 - z_3$.
However, it is difficult in general to cook up an architecture which is appropriate 
for arbitrary high-dimensional functions which have multiple discontinuities along nonlinear 
hyper-surfaces and highly nonlinear components, as will be introduced in the following examples. On the other hand, CCR is flexible and easy to implement,
and its basic components are well-understood. Furthermore, discontinuous activation functions are problematic for gradient-based optimization methods, 
which are commonly employed. 

\begin{figure}%[!htb]
%\begin{center}
%\centering
\includegraphics[width=5.27in, height=1.20in, keepaspectratio=false]{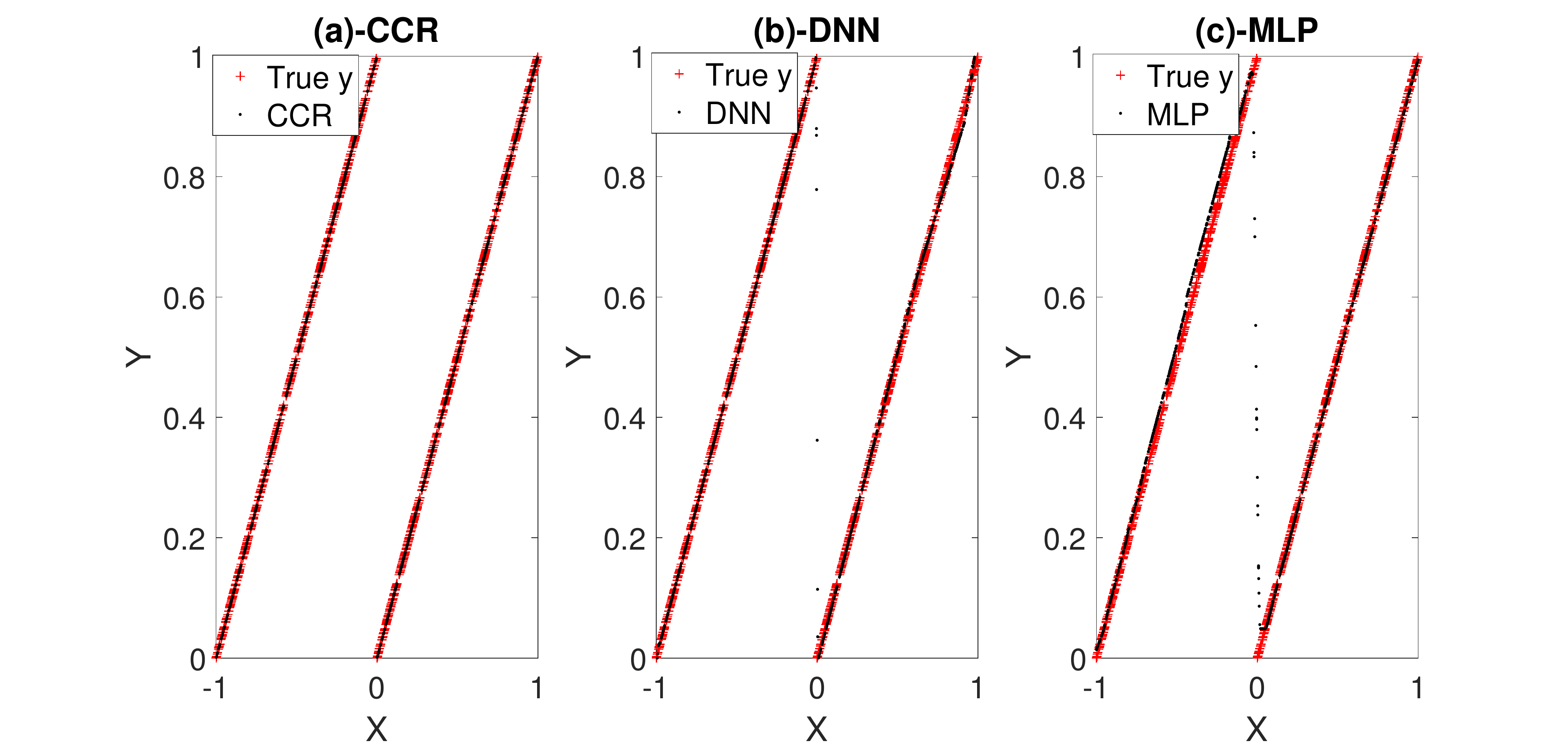}
%\center \textbf{Figure 2(iii):}
%\end{center}
\caption{The results of CCR (a), DNN (b), and MLP(c) as applied to numerical example 2 $f_2$.}
\label{fig:2} 
\end{figure}

To illustrate the benefit of CCR with simple MLPs
in comparison to alternative regression methods, 
we compare two direct regression methods on $f_2$.
The first is an MLP with a single hidden layer of 100 neurons
with ReLU activation functions, and a linear output.
The second is a deep neural network (DNN) with 
3 hidden layers of 200, 420, and 21 neurons each, with
ReLU activation functions, and linear output.
The results are presented in Figure \ref{fig:2}.
It is clear that neither of the simple regression methods 
are able to cleanly recover the discontinuity.

Now we consider slightly more complicated functions, 
following the recent work \cite{monterrubio2018posterior}.
%\ref{}. 
First, consider $\cX=[-4,10]$ and 
\begin{equation}\label{eq:f3}
f_3(x) = \left \{ \begin{array}{cc} 
\exp(-x^2/20) & x \leq 4 \\
1 & 4 < x \leq 6 \\
-1 & 6 < x \leq 8 \\
0 & x > 8 
\end{array} \right \} \, .
\end{equation}
Notice that the range of $f_3$ is very small in comparison to the domain $\cX$.
Therefore this is a prime example where 
the transformation \eqref{eq:scaling} needs to be utilized.
The method is illustrated on this example in Figure \ref{fig:one} row (c),
whose panels are the same as the rows above. %Figure \ref{fig:one}.
%A random forest method is used in this example for both regression and classification.
%For this example, we find that the MLP classification and regression method
%WHAT?? tends to favor constant functions on the individual clusters, 
%resulting in a piecewise constant reconstruction. 
%{\bf Therefore, a random forest method is used ???????}
%The function is plotted in Figure 3(a), 
%along with some noisy observations, 
%and the ultimate reconstruction. 
%Figure 3(b), 3(c) and 3(d) show, respectively, 
%the clustering result with our novel CCR method, the r-fit correlation line between CCR and the true signal and finally a histogram showing the dissimilarity between the signal recovered with CCR and the true signal

%\begin{figure}[!htb]
%\begin{center}
%%\centering
%\includegraphics[width=5.27in, height=2.95in, keepaspectratio=false]{nNUM3.eps}
%%\center \textbf{Figure 3:}
%%\label{fig:2a}} 
%\end{center}
%\caption{Numerical example 3. 
%The function is plotted in Panel (a), 
%along with the clustering values of the training data.
%Panel (b) shows prediction results on test data: 
%the final CCR machine output $f_r(x ; f_c(x))$, 
%the true $y(x)$, and the intermediate $f_c(x)$.
%Panel (c) shows a scatter plot of the true $y(x)$ and the CCR machine $f_r(x ; f_c(x))$,
%illustrating the correlation.
%Panel (d) shows a histogram of $f_r(x ; f_c(x))-y(x)$,
%illustrating the dissimilarity between the CCR reconstruction and the truth.}
%\label{fig:three}
%\end{figure}

\begin{figure}[!htb]
\begin{center}
%\centering
\includegraphics[width=5.27in, height=2.95in, keepaspectratio=false]{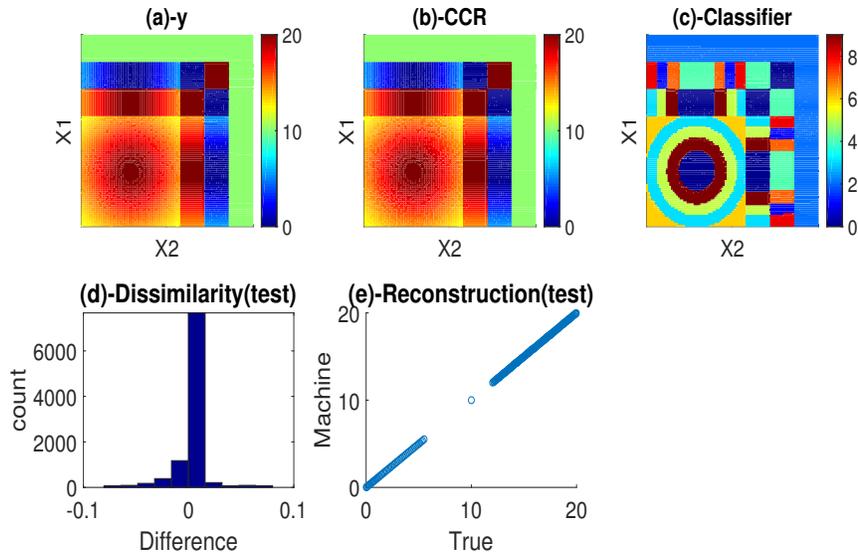}
%\center \textbf{Figure 4:}
%\label{fig:2a}} 
\end{center}
\caption{Numerical example 4. 
$f_4(x)$ is plotted in panel (a), %$\lambda(x)$ in panel (b) for the training data $x$,
and $f_r(x,f_c(x))$ and $f_c(x)$ are plotted in panels (b) and (c), respectively.
Panel (e) shows a scatter plot of the true $y(x)$ and the CCR machine $f_r(x ; f_c(x))$,
illustrating the correlation.
Panel (d) shows a histogram of $f_r(x ; f_c(x))-y(x)$,
illustrating the dissimilarity between the CCR reconstruction and the truth.}
%Panel (g) shows the elbow plot for this example.}
\label{fig:four}
\end{figure}

The next example is simple the product of the previous, 
%again as in Figure 3, 
so $\cX=[-4,10]^2$ and $f_4(x) = f_3(x_1)f_3(x_2)$.
The method is illustrated on this example in 
%The function is plotted in 
Figure \ref{fig:four}. 
%Again random forest classification and 
%regression method are employed in this example.
%along with some noisy observations, 
%and the ultimate reconstruction. 
%The method is illustrated on this example in Figure \ref{fig:three}.
%%whose panels are 
Here the information is the same as Figure \ref{fig:one}, but now
$f_4(x)$ is plotted in panel (a), %$\lambda(x)$ in panel (b) for the training data $x$,
and $f_r(x,f_c(x))$ and $f_c(x)$ are plotted in panels (b) and (c), respectively.
Panel (e) shows a scatter plot of the true $f_4(x)$ and the CCR machine $f_r(x ; f_c(x))$,
illustrating the correlation.
Panel (f) shows a histogram of $f_r(x ; f_c(x))-y(x)$,
illustrating the dissimilarity between the CCR reconstruction and the truth.
Panel (g) shows the elbow plot. Notice that there is no clear elbow. 
One might choose $L=4$ clusters, but this is not sufficient to avoid 
any cluster spanning a discontinuity.
From a visual inspection, $L=7$ is the minimum number of continuous components,
if we group components which meet at a point. In this case, 
sometimes we get a good set of clusters, but sometimes some clusters span 
a discontinuity.
If we let these each correspond to 2 distinct components, then $L=10$, 
which we find is sufficient to ensure no single cluster spans a discontinuity.
This is another illustration of the robustness of choosing extra clusters.
Interestingly, the clusters we find do not correspond to the 
continuous components at all, but rather (as in previous examples 2-3) 
partition the function primarily based on $y$-value contour level set intervals, 
crucially none of which span a discontinuity. 
%See Figure \ref{fig:clusters} for examples partitioning into $L=4$, and $L=7$ 
%clusters with k-means.

%The function is plotted in Figure 4(a) with the clustering groups plotted in Figure 4(b), 
%Figure 4(b) and 4(c) shows the signal recovered by CCR with the predicted classes. Figure 4(d) and 4(e) show, the r-fit correlation line between CCR and the true signal and a histogram showing the dissimilarity between the signal recovered with CCR and the true signal

%
%\begin{figure}[!htb]
%\begin{center}
%%\centering
%\includegraphics[width=5.27in, height=2.95in, keepaspectratio=false]{nNUM4.eps}
%%\center \textbf{Figure 4:}
%%\label{fig:2a}} 
%\end{center}
%\caption{Different clusters for numerical example 4 with $L=4$ (left) 
%and $L=7$ (middle and right).}
%\label{fig:clusters}
%\end{figure}

\subsection{Critical gradient model for tokamaks}\label{ssec:critgrad}

A tokamak is a device which uses magnetic fields to 
confine hot plasma in the shape of a torus. It is the leading 
candidate for production of controlled thermonuclear power, 
for use in a prospective future fusion reactor \cite{tokamakweb}.

One dimensional radial transport modeling \cite{park} using theory-based models such as GLF23 \cite{waltz1997gyro}, MMM95 \cite{bateman1998predicting}, and TGLF \cite{staebler2007theory} plays an essential role in interpreting experimental data and guiding new experiments for magnetically confined plasmas in tokamaks. Turbulent transport resulting from micro-instabilities have a strong nonlinear dependency on the temperature and density gradients. One of the key characteristics is a sharp increase of turbulent flux as the gradient of temperature increases beyond a certain critical value.
This leads to a highly nonlinear and discontinuous function of the inputs.

\begin{figure}%[!htb]
\centering
%\begin{minipage}[b][6cm]{0.5\linewidth}
%\subfloat[fu]
%\begin{tabular}{cc}
%\multirow[2]
%\end{minipage}%
%\begin{minipage}[t][6cm]{0.5\linewidth}
%\subfloat[it]{
\subfigure[Some slices of $f_5=\chi$.]
{\includegraphics[width=0.5 \columnwidth]{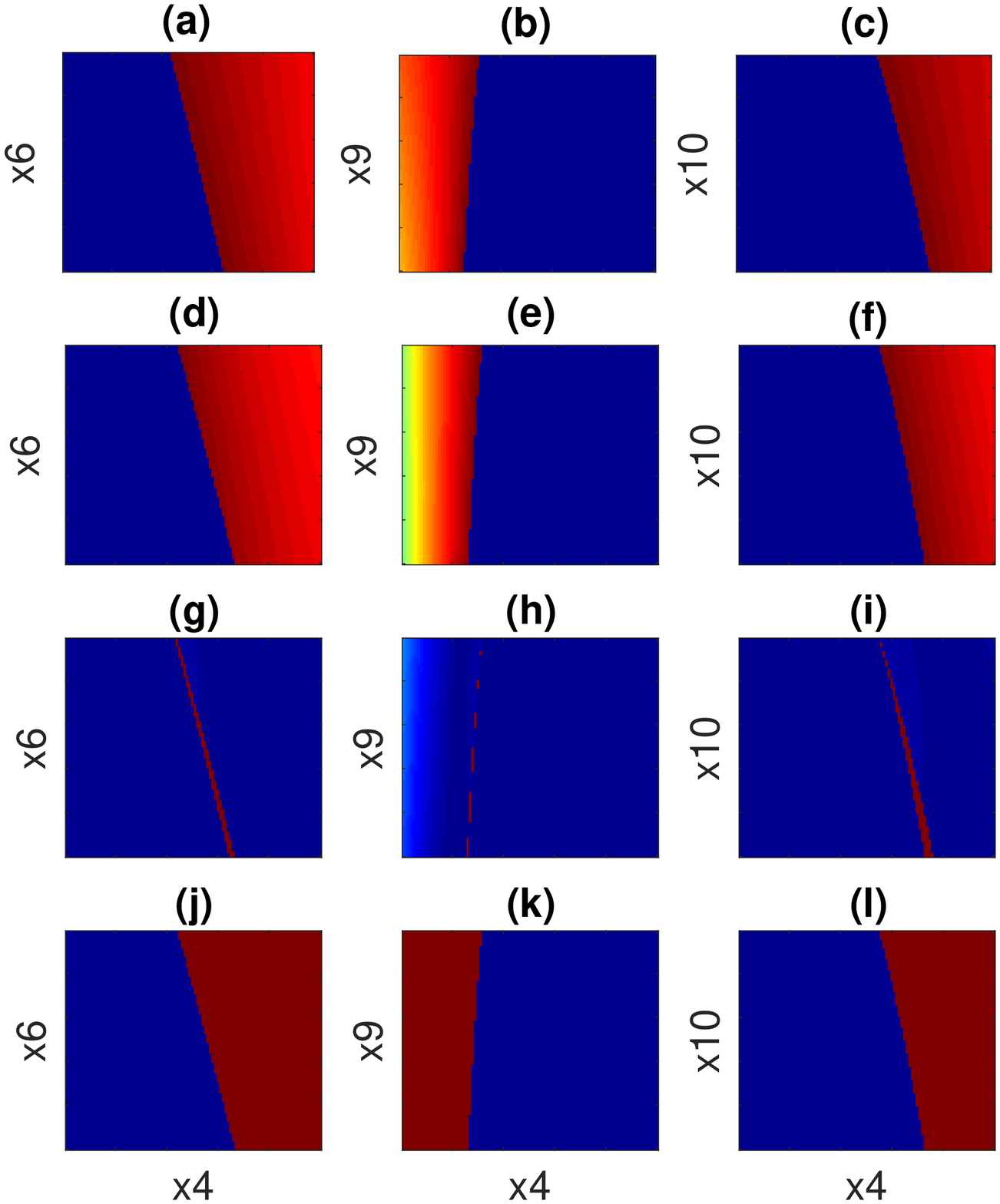}}
 %& 
 %\begin{tabular}{c}
 %\subfigure[b]{\includegraphics[width=0.45 \columnwidth]{5Tru}} %\\
%{\subfloat[C]
%& 
\subfigure[Input distribution marginals.]
{\includegraphics[width=0.45 \columnwidth, height=0.57\columnwidth]{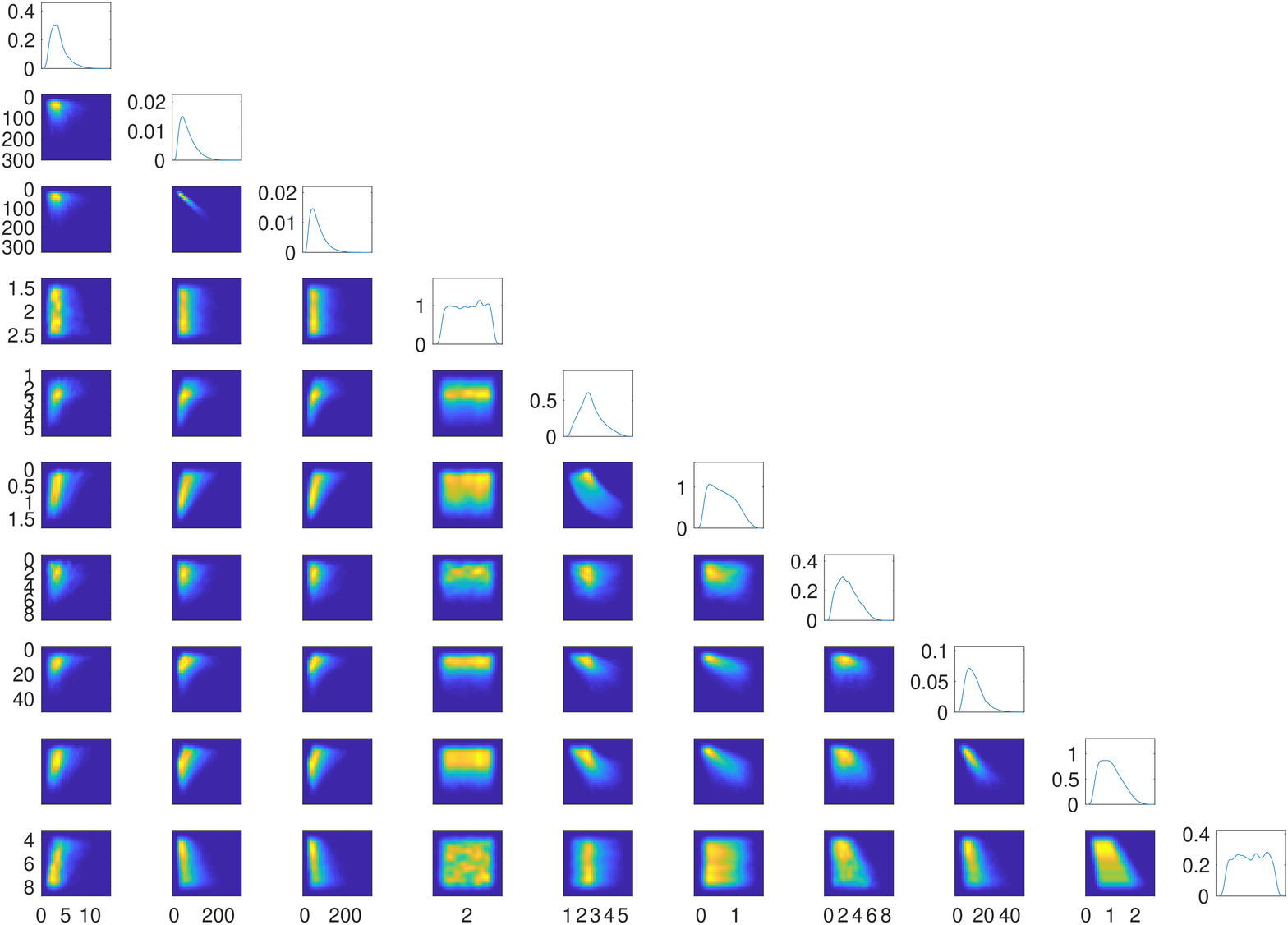}}
%\end{tabular}
%\end{tabular}
%\end{minipage}
%\end{center}
\caption{Numerical example 5: subfigure (a) 
shows some %single and 
two variable slices of the true function $\chi$ (a-c), 
the CCR machine output $f_r(x ; f_c(x))$ (d-f), 
the absolute difference $|\chi - f_r(x ; f_c(x))|$ (g-i), 
and the intermediate $f_c(x)$ (j-l), 
with remaining inputs set to the mean $\bbE(x_{\backslash ij})$, 
where $x_{\backslash ij}=(m_1,\dots, m_{j-1}, m_{j+1}, \dots 
m_{i-1},m_{i+1},\dots, m_{10})$. 
%The remaining 2 variable slices are plotted in 
Subfigure (b) shows the input data distribution marginals.}
\label{fig:5truth}
\end{figure}

Here we consider an analytic stiff transport model that describes turbulent ion energy transport in tokamak plasmas \cite{janeschitz20021}:
\begin{equation}\label{eq:chi}
\chi = S (R T'/T - (R T'/T)_{\rm crit})^\alpha 
H\left (\left |\frac{(R T'/T)}{(R T'/T)_{\rm crit}} \right | - 1\right) \, ,
\end{equation}
%?=S (?RT_i^'/T_i  -(RT_i^'/T_i)?_crit^IFPPL )^? H(|(RT_i^'/T_i)/?(RT_i^'/T_i)?_crit^IFSPPL |-1)
where $\chi$ %?_i 
is ion thermal diffusivity, $H(\cdot)$ is the Heaviside function, 
$R$ is the major radius, 
and $T'$ is the radial derivative of ion temperature. 
The normalized critical gradient 
%?(RT_i^'/T_i)?_crit^IFPPL 
$(R T'/T)_{\rm crit}$ of ion temperature is calculated using IFS/PPPL model \cite{kotschenreuther1995quantitative}, 
which is a nonlinear function of electron density ($n_e$), electron and ion temperatures ($T_e,T$), safety factor ($q$), magnetic shear ($\hat{s}$), effective charge ($Z_{\rm eff}$) 
and the normalized gradient of ion temperature $(RT'/T)$ and density $(Rn'/n)$. It is assumed that $S=1$ and $\alpha=1$.
Considering $(T, T')$ and $(n, n')$ as two input parameters
each, this gives a total of 10 inputs $x \in \bbR^{10}$.
The output \eqref{eq:chi} is $y \in \bbR_+$.
Basic primitive model inputs are chosen uniformly at 
random from a hypercube $\omega \in [0,1]^{17}$, 
which then give rise to realistic inputs $x \in \bbR^{10}$, 
which concentrate on a manifold in the ambient space.
See \cite{kotschenreuther1995quantitative} for details
of the model, and Figure \ref{fig:5truth} (b) for visualization
of the input distribution histogram.
It is difficult to visualize a function over $\bbR^{10}$, 
so we plot some slices in Figure \ref{fig:5truth} (a)
in order to get a sense of it.
This is now explained. Let $m=\bbE(x)$, 
where the expectation is with respect to the input distribution.
%For each $i = 1, \dots, 10$ and $j=1, \dots, 10$,
%the diagonal plots of Figure \ref{fig:5truth} correspond to $\chi(x)$ as $x$ 
%varies over a $1d$ grid $(m_1,\dots, m_{i-1},x_i, m_{i+1},\dots, m_{10})$, 
%and the off-diagonal plots correspond to 
For $i=4$ and $j=6, 9, 10$ (in rows 1, 2, 3, respectively), 
$\chi(x)$ is plotted 
as $x$ varies over a $2d$ grid $(m_1,\dots, m_{j-1}, x_j, m_{j+1}, \dots 
m_{i-1},x_i, m_{i+1},\dots, m_{10})$.

\begin{figure}[!htb]
\begin{center}
%\centering
\includegraphics[width=5.27in, height=2.95in, keepaspectratio=false]
{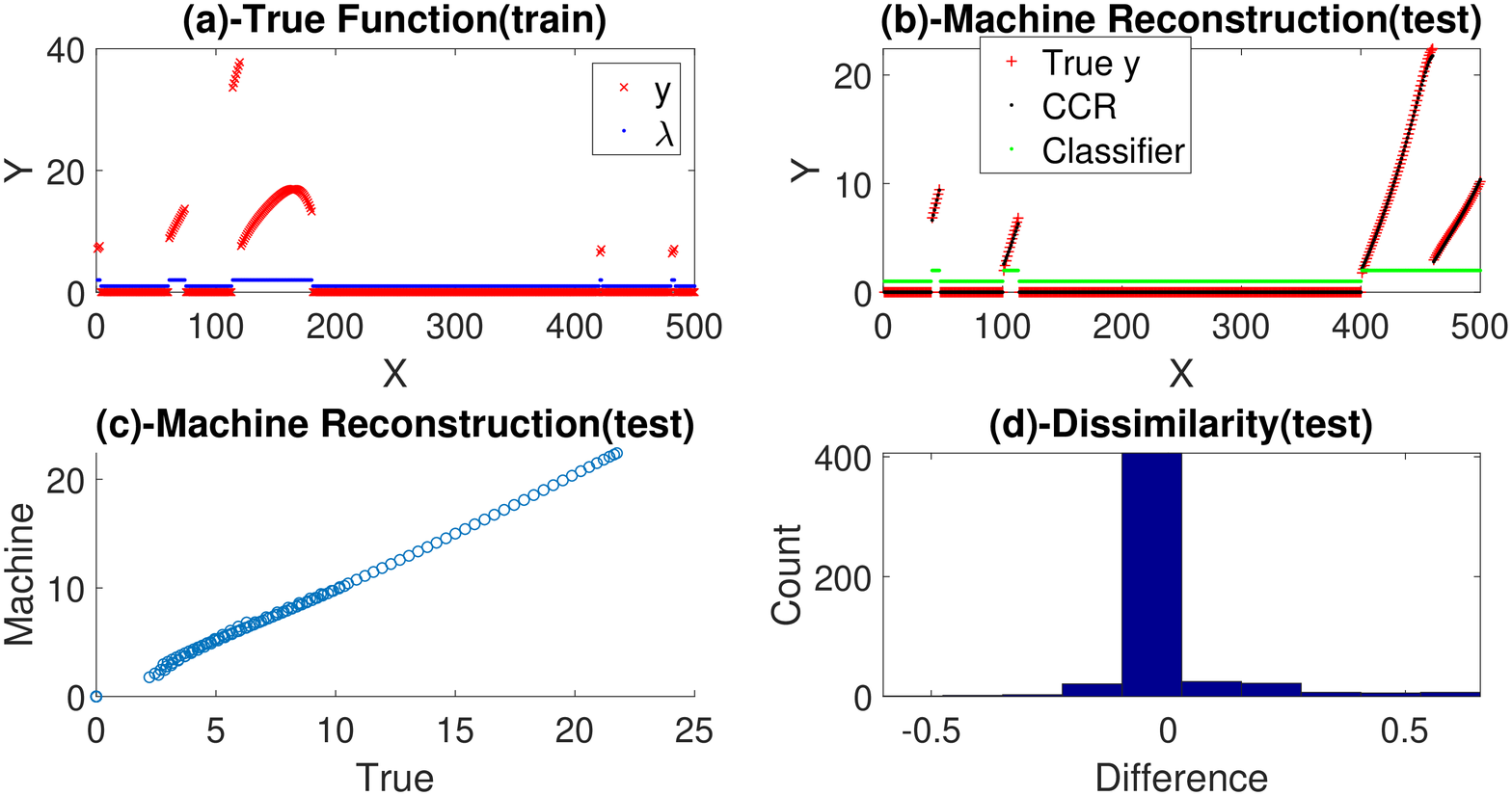}
%\center \textbf{Figure 5:}
%\label{fig:2a}} 
\end{center}
\caption{Numerical example 5. 
The first 500 (random) training data output values are plotted in Panel (a), 
along with the clustering values of the training data.
Panel (b) shows prediction results on test data: 
the final CCR machine output $f_r(x ; f_c(x))$, 
the true $y(x)$, and the intermediate $f_c(x)$.
Panel (c) shows a scatter plot of the true $y(x)$ and the CCR machine $f_r(x ; f_c(x))$,
illustrating the correlation.
Panel (d) shows a histogram of $f_r(x ; f_c(x))-y(x)$,
illustrating the dissimilarity between the CCR reconstruction and the truth.}
\label{fig:five}
\end{figure}

%{\bf CRITICAL GRADIENT MODEL HERE}
%\cite{kotschenreuther1995quantitative}
The CCR method is illustrated on this example in 
Figure \ref{fig:five}. Here $N_{\rm train}=300000$ training data points are used, 
and $N_{\rm test}=500$ test data points.
%Figure \ref{fig:five}. Here $N=300000$ training data points are used, 
%and $M=500$ test data points.
%whose panels are the same as Figure \ref{fig:one}.

%\begin{figure}[!htb]
%\begin{center}
%%\centering
%\includegraphics[width=1\columnwidth]{data_input_marginals}
%%\center \textbf{Figure 5:}
%%\label{fig:2a}} 
%\end{center}
%\caption{Numerical example 5: single and two variable
%marginal histograms of inputs.}
%\label{fig:5histos}
%\end{figure}

%The function is plotted in Figure 5(a), 
%along with some noisy observations, 
%and the ultimate reconstruction.
%Figure 5(b), 5(c) and 5(d) show, respectively, 
%the clustering result with our novel CCR method, the r-fit correlation line between CCR and the true signal and finally a histogram showing the dissimilarity between the signal recovered with CCR and the true signal

\subsection{Accuracy and active learning extensions}

\begin{table}
\begin{tabular}{|p{0.8in}|p{0.8in}|p{0.8in}|p{0.8in}|p{0.8in}|p{0.8in}|} 
\hline 
\textbf{Accuracy} & \textbf{Example 1} & \textbf{Example 2} & \textbf{Example 3} & \textbf{Example 4} & \textbf{Example 5} \\ \hline 
\textbf{L2 (\%)} & 0.9934 & 0.9961 & 0.9964 & 0.9978 & 0.9825 \\ \hline 
\textbf{R2 (\%)} & 0.9978 & 0.9967 & 0.9987 & 0.9983 & 0.9845 \\ \hline 
\end{tabular}
\caption{L2 and R2 comparison for the 5 numerical examples.}
\label{tab:one}\end{table}
The accuracy of the methods on test data is presented in Table \ref{tab:one}.
%In particular, {\bf N data points are used for training, 
%and then M test data points are used to compute the following measures of accuracy}
%\begin{equation}\label{eq:l2}
%L2 = 1 - \sqrt{\frac{\sum_{i=N+1}^{N+M} |f_r(x_i;f_c(x_i)) - y_i|^2}{\sum_{i=N+1}^{N+M} |y_i|^2}} \, ,
%\end{equation}
%and
%\begin{equation}\label{eq:l2}
%R2 = 1 - \frac{\sum_{i=N+1}^{N+M} |f_r(x_i;f_c(x_i)) - y_i|^2}
%{\sum_{i=N+1}^{N+M} |y_i - \bar{y}|^2} \, ,
%\end{equation}
%where $\bar{y} = \frac1{M} \sum_{i=N+1}^{N+M} y_i$.
In particular, 
\begin{equation}\label{eq:l2}
L2 = 1 - \sqrt{\frac{\sum_{i=1}^{N} |f_r(x_i;f_c(x_i)) - y_i|^2}{\sum_{i=1}^{N} |y_i|^2}} \, ,
\end{equation}
and
\begin{equation}\label{eq:l2}
R2 = 1 - \frac{\sum_{i=1}^{N} |f_r(x_i;f_c(x_i)) - y_i|^2}
{\sum_{i=1}^{N} |y_i - \bar{y}|^2} \, ,
\end{equation}
where $\bar{y} = \frac1{N} \sum_{i=1}^{N} y_i$. 
For example 5, the data used to compute the error is out-of-sample
testing data. For the grid-based examples, it is the training data. % on the grid.
\begin{table}
\begin{tabular}{|p{0.8in}|p{0.8in}|p{0.8in}|p{0.8in}|p{0.8in}|p{0.8in}|} \hline 
%\textbf{Error} 
& \textbf{Active}% learning-CCR} 
& \textbf{Passive}% Learning-CCR}
\\ \hline 
\textbf{Error} & 0.1 & 0.15 \\ \hline 
\textbf{$N$} & 150 & 1000  \\ \hline 
\end{tabular}
\caption{Error attainment with set of sample points for active learning 
with Example 2 and strategy 1a: $N_{\rm res}=1000$ 
and all points are used for passive learning, 
while only $n=150$ points are used for active.}
\label{tab:two}
\end{table}
Table \ref{tab:two} illustrates active learning with active strategy 1a
from Section \ref{sec:active}.

%\noindent 
%\subsubsection{Active learning adaptation}

%\noindent  For active learning, we utilise pool sampling. Active Learning is a special case of Machine Learning in which a learning algorithm is able to interactively query the user to obtain the desired outputs at new data points.The process of subsetting the data is done with an Active Learner which is going to learn based on a strategy, which training subsets are appropriate for maximising the accuracy of our model.
%\begin{enumerate}
%  \item Random sampling: the data points are sampled at random
%  \item Uncertainty sampling: we select the points whose class we are most uncertain about.
%  \item Entropy sampling: we choose the points whose class probability have the largest entropy
%  \item Margin sampling: we choose the points for whom the difference between the most and second most likely classes are the smallest.
%\end{enumerate}
%
%In this paper we choose uncertainty sampling as our querying strategy for sampling new points.

\section{Discussion}\label{sec:conclusions}

In this work we present a simple general framework
for using the existing arsenal of fundamental machine learning tools 
in order to solve the challenging problem of reconstructing a surrogate
model, or machine, for approximating highly nonlinear and discontinuous 
functions over high dimensional spaces. We have illustrated the 
power and accuracy of the method on various examples.
It is notable that the method admits a huge amount of flexibility, 
as any combination of clustering, classification, 
and regression models will work. 
It can also be wrapped around existing deep learning architectures, 
which typically already combine elements of the latter 2.
Therefore, it is reasonable to call this \emph{super deep learning}.
The cost of the regression stage may be high if there are many classes/clusters,
since there is one regression per class, but this stage is also embarrassingly parallel.

%{\bf and compared to the state of the art ???}.

\vspace{10pt}
\noindent 
{{\bf Acknowledgements}: This work is supported by the U.S. Department of Energy, Office of Science, Office of Advanced Scientific Computing Research (ASCR) Scientific Discovery through Advanced Computing (SciDAC) project on Advanced Tokamak Modelling (AToM), 
under field work proposal number ERKJ123.}

\bibliographystyle{plain} 
\bibliography{references}

\end{document}